\documentclass[11pt]{article}

\usepackage[preprint]{acl}

\usepackage{times}
\usepackage{latexsym}

\usepackage[T1]{fontenc}

\usepackage[utf8]{inputenc}

\usepackage{microtype}

\usepackage{inconsolata}

\usepackage{graphicx}

\usepackage{multirow}
\usepackage{amsmath}
\usepackage{booktabs} 
\usepackage{graphicx}

\title{Towards Compositional Generalization of LLMs via Skill Taxonomy Guided Data Synthesis}

\author{
  \textbf{Yifan Wei\textsuperscript{1}},
  \textbf{Li Du\textsuperscript{2}\thanks{Corresponding Authors.}},
  \textbf{Xiaoyan Yu\textsuperscript{3}},
  \textbf{Yang Feng\textsuperscript{4}},
  \textbf{Angsheng Li\textsuperscript{1}\footnotemark[1]}
\\
\\
  \textsuperscript{1}State Key Laboratory of Complex \& Critical Software Environment, Beihang University \\
  \textsuperscript{2}Beijing Academy of Artificial Intelligence,
  \textsuperscript{3}Beijing Institute of Technology, \\
  \textsuperscript{4}Institute of Computing Technology, CAS
\\
  \small{
    \href{mailto:weiyifan@buaa.edu.cn}{weiyifan@buaa.edu.cn}, \href{mailto:duli@baai.ac.cn}{duli@baai.ac.cn}, \href{mailto:angsheng@buaa.edu.cn}{angsheng@buaa.edu.cn}
  }
}

\begin{document}
\maketitle

\begin{abstract}
Large Language Models (LLMs) and agent-based systems often struggle with compositional generalization due to a data bottleneck in which complex skill combinations follow a long-tailed, power-law distribution, limiting both instruction-following performance and generalization in agent-centric tasks.
To address this challenge, we propose STEPS, a \textbf{S}kill \textbf{T}axonomy–guided \textbf{E}ntropy-based \textbf{P}ost-training data \textbf{S}ynthesis
framework for generating compositionally challenging data. STEPS explicitly targets compositional generalization by uncovering latent relationships among skills and organizing them into an interpretable, hierarchical skill taxonomy using structural information theory.
Building on this taxonomy, we formulate data synthesis as a constrained information maximization problem, selecting skill combinations that maximize marginal structural information within the hierarchy while preserving semantic coherence.
Experiments on challenging instruction-following benchmarks show that STEPS outperforms existing data synthesis baselines, while also yielding improved compositional generalization in downstream agent-based evaluations. 
The code and data for our methods and experiments are available at \href{https://github.com/weiyifan1023/STEPS}{https://github.com/STEPS}.
\end{abstract}

\section{Introduction}

The rapid scaling of Large Language Models (LLMs) has led to impressive gains across language understanding and generation tasks. However, both LLMs and agent-based systems continue to struggle with compositional generalization—the ability to flexibly recombine learned skills into novel configurations—particularly in complex instruction-following and agent-centric settings \citep{lake2018generalization, okawa2023compositional}.
A key obstacle is a fundamental data bottleneck: while individual atomic skills are abundantly represented in training corpora, complex skill combinations follow a long-tailed, power-law distribution \citep{clauset2009power}. This imbalance severely limits coverage of compositionally challenging scenarios, leading to sharp performance degradation when multiple skills must be coordinated, as evidenced by evaluations such as SKILL-MIX \citep{kudo2023deep, yu2024skillmix, zhao2024can}.

To alleviate this data bottleneck, prior work has mainly explored data-centric strategies such as data mixture optimization and pedagogical sequencing, which reweight training samples or adjust learning order to improve sample efficiency \citep{ge2024bimix, wu2024mixture, chen2023skill, Zhao_Du_Ju_Wu_Pan_2025, hu2025step}. While effective at better utilizing existing data, these approaches primarily operate at the level of individual skills, and therefore do not fundamentally address the scarcity of compositionally challenging examples involving multiple interacting skills.
Complementary efforts synthesize such data through stochastic or heuristic skill mixing \citep{chen-etal-2024-skills, kaur2025instructskillmix}, but typically ignore the underlying structural and hierarchical relationships among skills, resulting in semantically incoherent compositions and inefficient exploration of the combinatorial space. 
From a broader perspective, existing methods lack an explicit, principled formulation of skill composition in the form of a unified taxonomy. Thus their ability to support systematic data synthesis is limited. This gap motivates the need for structure-aware synthesis approaches that explicitly model skill dependencies and compositions.

In this paper, we propose STEPS, a \textbf{S}kill \textbf{T}axonomy–guided \textbf{E}ntropy-based \textbf{P}ost-training data \textbf{S}ynthesis framework for addressing compositional data scarcity in the complex instruction following and agent-based systems.
As shown in Figure~\ref{fig:main}~(a), STEPS first induces a hierarchical skill dependency taxonomy by constructing a skill co-occurrence graph and recursively merging groups of skill nodes with dense intra-group connections and sparse inter-group connections. This process is guided by minimizing structural entropy \citep{li2016structural,li2024science}, which favors hierarchies in which high-weight co-occurrence edges are largely explained within local groups rather than across unrelated ones, yielding a compact and interpretable dependency structure over skills.
Then as illustrated in Figure~\ref{fig:main}~(b), building on this taxonomy, STEPS then synthesizes new data by recursively composing skill nodes under an objective of maximizing marginal structural entropy gain. Instead of sampling combinations at random, this strategy prioritizes skill compositions that introduce the largest amount of new structural information while remaining coherent with the induced hierarchy. As a result, STEPS efficiently explores the combinatorial skill space and generates compositionally challenging data that better supports generalization.

Experiments on several challenging instruction following benchmarks and on agent scenario demonstrate that our framework significantly enhances the model performance, providing a principled solution for overcoming data sparsity and advancing the compositional capabilities of LLMs.
\section{Preliminary}
\label{sec:pre}

\subsection{Problem Formalization}
Large Language Models (LLMs) are increasingly required to solve tasks that involve the coordinated use of multiple functional capabilities, which we refer to as \emph{skills}. Formally, a skill $s \in \mathcal{S}$ denotes an atomic functional ability required to execute an instruction, such as logical reasoning, mathematical calculation, or code debugging.
An instruction can therefore be characterized by a set of jointly involved skills, which we represent as a $k$-tuple $X = \{x_1, x_2, \dots, x_k\}$, where $k$ reflects the \emph{compositional complexity} of the task. Larger values of $k$ correspond to more challenging instructions that require integrating multiple interacting skills.

Our goal is to improve the compositional generalization of LLMs and agent-based systems by synthesizing training data that systematically covers informative and challenging skill combinations beyond those frequently observed in corpora.

\subsection{Skill Taxonomy of LLMs}

The capabilities of LLMs are not independent: complex skills are often built upon foundational ones, and certain skills frequently co-occur to solve intricate tasks. Motivated by this observation, we model the skill space as a \emph{skill taxonomy} $\mathcal{T}$, where each leaf node corresponds to an atomic skill $s_i$, each internal node represents a coherent group of closely related skills, and edges encode hierarchical dependency relationships.

Rather than treating the skill space as a flat set, we seek a structured abstraction that exposes its hierarchical organization. Specifically, we aim to induce a \emph{skill taxonomy} that groups closely related skills together while separating weakly related ones across different levels. This taxonomy provides a compact and interpretable representation of the combinatorial skill space, and serves as the structural foundation for selecting and synthesizing skill compositions.
From this perspective, data synthesis reduces to selecting skill combinations from the taxonomy: combinations confined to a narrow region tend to be redundant, while those connecting different yet related regions are more informative. This motivates a synthesis strategy that explicitly leverages the structure of the skill taxonomy when generating compositional training data.

\begin{figure*}
    \centering
    \includegraphics[width=0.9\linewidth]{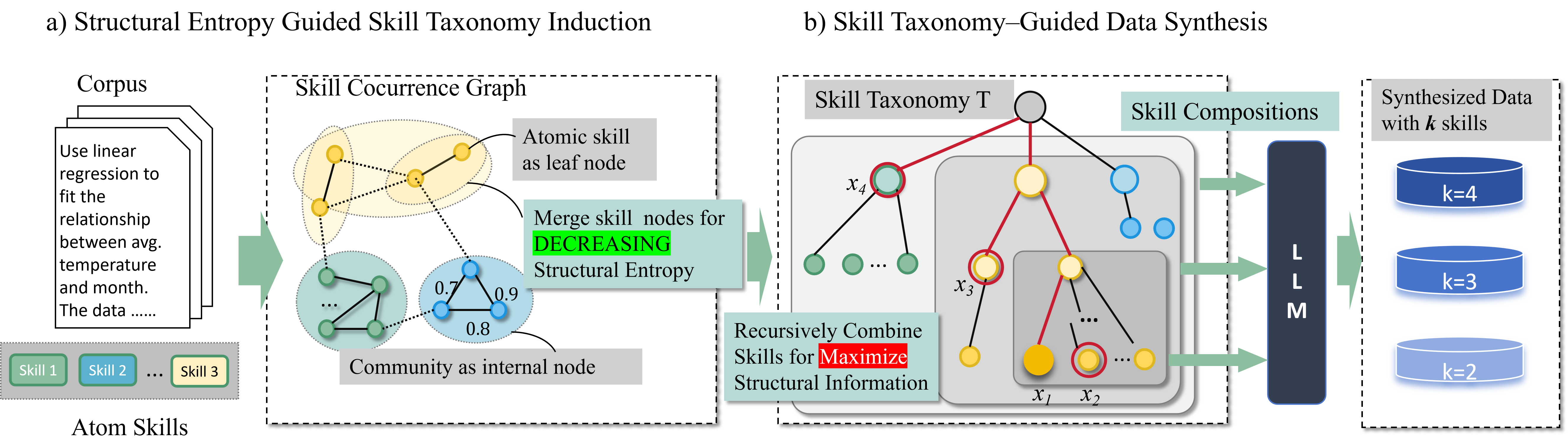}
    \caption{Illustration of the STEPS framework.}
    \label{fig:main}
\end{figure*}

\section{Methodology}
\label{sec:method}

Figure~\ref{fig:main} demonstrates the two-stage framework for taxonomy-guided data synthesis. We first induce a hierarchical skill taxonomy that captures dependency and compositional relationships among skills. Based on this taxonomy, we then synthesize training data by selecting skill combinations that maximize structural information gain under hierarchical constraints, yielding compositionally challenging yet learnable examples.

\subsection{Structural Entropy Guided Skill Taxonomy Induction}
\label{sec: skill taxonomy}
As shown in Figure~\ref{fig:main}~(a), to induce a skill taxonomy, we begin by modeling the relationships among skills through their empirical co-occurrence patterns. Following recent studies \citep{Zhao_Du_Ju_Wu_Pan_2025, li2025infinity, du2025scaling}, we annotate each instruction with a set of skill tags and construct a skill co-occurrence graph $G=(V,E)$, where each node $v\in V$ corresponds to an atomic skill and each edge weight reflects the probability of two skills co-occur in one real data.

While co-occurrence statistics capture local relationships, they do not directly reveal the global organization of the skill space, which is often sparse and long-tailed \citep{barabasi1999emergence, broder2000graph, clauset2009power}. To uncover higher-level structure, we seek to hierarchically merge closely related skills into more abstract skill groups, forming a skill dependency taxonomy. This process can be viewed as recursively grouping skill nodes with dense intra-group connections and sparse inter-group connections, yielding progressively higher-level abstractions.

We formalize this objective using structural entropy \citep{li2016structural,wei2025structural}, which provides a principled criterion for identifying such groupings. Intuitively, structural entropy is low when high-weight co-occurrence edges are largely explained within local groups. Minimizing structural entropy therefore favors hierarchies in which strongly related skills are clustered together, while weakly related skills are separated across levels.

Formally, given a taxonomy $\mathcal{T}$ represented as a tree structure, we extend structural entropy from communities to individual skill nodes. Let $\gamma$ denote the leaf node in $\mathcal{T}$ corresponding to skill $v$, and $\lambda$ the root of the tree. The structural entropy of a skill $v$, denoted $S_e(v)$, is defined as the cumulative entropy contribution along the path from $\gamma$ to $\lambda$:
\begin{equation}
\begin{aligned}
S_{e}(v)
&= \mathcal{H}^{\mathcal{T}}(G; v)
= \mathcal{H}^{\mathcal{T}}(G; [\gamma, \lambda)) \\
&= - \sum_{\lambda \subset \alpha \subseteq \gamma}
\frac{g(\alpha)}{vol(G)} \log \frac{vol(\alpha)}{vol(\alpha^-)} ,
\end{aligned}
\end{equation}
where $\alpha$ ranges over the community nodes on the path from $\gamma$ to $\lambda$, $g(\alpha)$ denotes the boundary weight of $\alpha$, and $vol(\cdot)$ its volume. A higher $S_e(v)$ indicates that the skill lies in a less cohesive region of the taxonomy, suggesting greater potential for forming diverse and informative compositions.

We induce the skill taxonomy using a bottom-up agglomerative procedure. Starting from the skill co-occurrence graph $G=(V,E)$, we initialize each atomic skill $v\in V$ as a singleton community, which corresponds to a leaf in the taxonomy. At each step, we consider merging a pair of communities and choose the merge that yields the largest decrease in structural entropy. The merged community is added as a new internal node whose children are the two merged communities. We repeat this process until all skills are merged into a single community, which becomes the root of the taxonomy. This procedure yields a hierarchical coding tree $T$ that groups strongly co-occurring skills at lower levels while separating weakly related ones at higher levels.


\subsection{Skill Combination via Structural Information Maximization}
\label{sec:info_max}
Given the induced skill taxonomy $T$, our next goal is to synthesize data with skill combinations that can rapidly increase the compositional generalizability of models. Rather than treating all combinations equally, given the induced skill taxonomy $\mathcal{T}$, our objective is to identify skill combinations that provide the largest structural information gain for training. Intuitively, not all $k$-skill combinations are equally informative: combinations that remain within a small, well-explored region of the taxonomy tend to be redundant, whereas those that introduce new structural relationships across different regions can yield greater learning benefit.

To formalize this notion, we seek a principled measure of the incremental structural information contributed by adding a new skill to an existing set. Importantly, the informational value of a skill depends on which skills have already been selected, as overlapping skills may share structural context in the taxonomy. This dependency makes it insufficient to evaluate skill combinations using independent or additive scores.

To address this, we utilize the chain rule of entropy, which allows the joint structural entropy of a skill set to be decomposed into a sequence of conditional terms. This decomposition enables us to quantify the marginal structural information gain contributed by each skill relative to the previously selected ones, providing a natural objective for constructing informative skill combinations.

For a $k$-tuple of skills $X=\{x_1,x_2,\dots,x_k\}$, we decompose its joint structural entropy into a sum of conditional terms. Specifically, the conditional structural entropy
$\mathcal{H}^T(G; x_i \mid X_{<i})$ measures the marginal structural information contributed by skill $x_i$ relative to the previously selected set $X_{<i}=\{x_1,\dots,x_{i-1}\}$.
Let $\delta$ denote the least common ancestor (LCA) of ${x_i}\cup X_{<i}$ in the taxonomy $\mathcal{T}$, and let $\gamma$ be the leaf node corresponding to $x_i$. The conditional structural entropy is defined as:
\begin{equation}
\label{equ:conditional_entropy}
\mathcal{H}^\mathcal{T}(G; x_i \mid X_{<i})
= - \sum_{\delta \subset \alpha \subseteq \gamma}
\frac{g(\alpha)}{vol(G)} \log \frac{vol(\alpha)}{vol(\alpha^-)} ,
\end{equation}
which captures only the novel structural information introduced by $x_i$ beyond what is already explained by $X_{<i}$.

We construct a skill combination $X$ using a greedy sequential procedure. Starting from an initial skill $x_1$, we iteratively select:
\begin{equation}
\label{equ:step2}
x_{i+1} = \arg\max_{v \in V \setminus X_{<i}}
\mathcal{H}^\mathcal{T}(G; v \mid X_{<i}) .
\end{equation}
Repeating this process yields a $k$-tuple that maximizes joint structural entropy, prioritizing skill combinations that are diverse and compositionally challenging.

\subsection{Entropy-Guided Recursive Synthesis}
\label{sec:recursive_synthesis}
Maximizing structural entropy provides a principled way to identify informative skill combinations. However, unconstrained maximization may result in combinations that are overly disparate or semantically irrational. Effective composition requires balancing diversity with reasonability, i.e, achieving a ``sweet spot''.

To achieve this balance, we introduce a recursive, taxonomy-guided search strategy, as shown in Figure~\ref{fig:main}~(b). Instead of performing global maximization over all skills, the selection of each subsequent skill is first restricted to the local sub-trees in the taxonomy. The search space is then gradually expanded to higher-level parent communities. This bottom-up expansion ensures  selected skills are structurally diverse while remaining within a coherent hierarchical context.

Once a skill combination is selected, we synthesize training data based on the identified skills with corresponding reference examples. Leveraging the strong meta-learning capability of LLMs, we prompt the model to generate new instructions and solutions that jointly exhibit the target skills. This process yields challenging training instances with compositions of skills for the generalization of models. More details are provided in the Appendix.

\section{Experiments}
\label{sec:experiments}

We evaluate our framework STEPS through extensive experiments on multiple instruction-following benchmarks. Our study aims to address the following research questions:
\textbf{RQ1}: Can our framework effectively enhance the compositional generalization of existing LLMs?
\textbf{RQ2}: What is the impact of $k$-tuple size on the model’s generalization capabilities?
\textbf{RQ3}: Can the hierarchical structure of STEPS effectively guide the learning process to enhance compositional generalization?
\textbf{RQ4}: Does the "sweet spot" constraint provide a more effective scope for skill learning?
\textbf{RQ5}: Can the STEPS method be used to construct agent data?

\subsection{Experimental Settings}

\noindent \textbf{Benchmarks and Metrics.} We evaluate model performance on three benchmarks:  MT-Bench \cite{zheng2023judging}, AlpacaEval 2.0 \cite{dubois2024lengthcontrolled}, and WildBench \cite{lin2025wildbench}. 
For AlpacaEval 2.0, we report the Length-Controlled Win Rate (LC WR) against a reference model to mitigate length bias. For MT-Bench, we report the average score of the responses of our model graded by a judge model. 
For WildBench, we utilize the WB-Score to evaluate the model's proficiency in following complex instructions. This metric provides a weighted average of task-wise performance evaluated by GPT-4o \citep{hurst2024gpt}.

\noindent \textbf{Baselines.} We select three mainstream open-source LLMs as base and instruct models: Qwen-2.5-7B, Llama-3-8B, and Mistral-7B-v0.3 \citep{jiang2023clip,team2024qwen2,grattafiori2024llama}. We compare our framework against two representative synthetic data methods: (1) Alpaca52k \cite{alpaca}, a large-scale dataset (52K samples) designed for general instruction following; and (2) Instruct-SkillMix \cite{kaur2025instructskillmix}, a state-of-the-art compositional method that utilizes random skill pairing. To emphasize the role of data quality, both our method and Instruct-SkillMix are fine-tuned on a compact set of 4K synthetic examples.

\subsection{Main Results (RQ1)}
\label{sec:main_results}

\begin{table}[hbt!]
\centering
\caption{Comparison of Performance on AlpacaEval 2.0 and MT-Bench for Different Models}
\label{tab:model_instruction_tuning_en}
\resizebox{\linewidth}{!}{
\begin{tabular}{l c c c}
\toprule
\multirow{2}{*}{Model} & \multirow{2}{*}{Data} & \textbf{AlpacaEval 2.0} & \textbf{MT-Bench} \\
& & LC WR (\%) & Score \\
\midrule
\multicolumn{4}{l}{\textbf{Mistral-7B-v0.3 \quad Base/Instruct}} \\
\cmidrule(lr){1-4}
Original & - & 0.19 / 22.88 & 3.44 / 6.63 \\
Alpaca52k & 52K & 4.57 / 5.38 & 3.79 / 4.03 \\
Instruct-SkillMix & 4K & 17.17 / 22.63 & 5.19 / 5.95 \\
STEPS & 4K & 17.60 / \textbf{24.42} & 5.85 / 6.23 \\
\midrule
\multicolumn{4}{l}{\textbf{Llama-3-8B \quad Base/Instruct}} \\
\cmidrule(lr){1-4}
Original & - & 0.20 / 27.85 & 2.04 / 6.94 \\
Alpaca52k & 52K & 2.09 / 9.37 & 4.13 / 5.04 \\
Instruct-SkillMix & 4K & 0.47 / 30.55 & 4.55 / 7.15 \\
STEPS & 4K & 4.65 / \textbf{35.26} & 4.66 / \textbf{7.35} \\
\bottomrule
\end{tabular}
}
\end{table}

\begin{table*}[hbt!]
\centering
\small
\caption{Performance comparison on WildBench. We report both Base and Instruct versions (Base / Instruct) for each model. The results across five task categories and the overall WB-Score demonstrate that our framework consistently outperforms original models and competitive synthesis baselines.}
\label{tab:combined_model_performance}

\begin{tabular}{lcccccc}
\toprule
\textbf{Model} & \textbf{Info Seek} & \textbf{Creative} & \textbf{Code \& Debug} & \textbf{Reason \& Plan} & \textbf{Math \& Data} & \textbf{WB-Score} \\
\midrule
DeepSeek-V2-Chat  & 52.7 & 56.4 & 55.0 & 54.8 & 51.4 & 54.0 \\
GPT-4-turbo & 57.2 & 58.7 & 55.1 & 56.2 & 51.0 & 55.2 \\
GPT-4o  & 58.6 & 59.1 & 60.5 & 60.2 & 57.3 & 59.3 \\
\midrule
\multicolumn{7}{c}{\textbf{Qwen-2.5-7B   \quad  Base/Instruct}} \\
\cmidrule(lr){1-7}
- Original  & 21.5 / 47.2 & 6.5 / 44.5 & 5.3 / 44.7 & 12.5 / 48.9 & 5.5 / 44.4 & 10.7 / 45.4 \\
- Alpaca52k & -6.3 / -13.4 & -5.4 / -14.3 & -2.0 / -9.1 & 3.3 / -6.5 & -7.7 / -14.1 & -3.8 / -11.4 \\
- Instruct-SkillMix & 35.3 / 41.9 & 33.8 / 36.6 & 27.9 / 41.7 & 36.1 / 40.8 & 31.7 / 37.0 & 32.8 / 39.6 \\
- STEPS & 39.3 / 44.0 & 34.5 / 45.1 & 31.1 / 49.1 & 39.5 / 48.3 & 34.0 / 45.5 & 35.1 / 45.7 \\
\midrule
\multicolumn{7}{c}{\textbf{Llama-3-8B   \quad  Base/Instruct}} \\
\cmidrule(lr){1-7}
- Original  & -27.7 / 37.9 & -30.4 / 43.0 & -23.8 / 26.0 & -17.1 / 35.8 & -29.8 / 22.4 & -25.7 / 34.2 \\
- Alpaca52k & 0.5 / 15.9 & -18.6 / -10.4 & -6.5 / 4.1 & 0.4 / 11.3 & -16.6 / -9.1 & -6.1 / 5.0 \\
- Instruct-SkillMix & 22.4 / 39.1 & 13.8 / 20.6 & 23.9 / 32.5 & 26.4 / 37.7 & 9.7 / 15.7 & 19.7 / 30.8 \\
- STEPS & 30.6 / 46.1 & 7.7 / 25.2 & 26.8 / 37.9 & 30.6 / 42.9 & 12.3 / 18.9 & 23.0 / 36.3 \\
\midrule
\multicolumn{7}{c}{\textbf{Mistral-7B-v0.3 \quad  Base/Instruct}} \\
\cmidrule(lr){1-7}
- Original  & -26.1 / 39.5 & -28.1 / 20.7 & -21.2 / 29.3 & -9.3 / 33.7 & -29.8 / 8.5 & -22.7 / 28.6 \\
- Alpaca52k & -14.2 / -13.6 & -24.8 / -26.6 & -13.7 / -13.2 & -5.1 / -8.2 & -20.0 / -12.5 & -14.0 / -14.8 \\
- Instruct-SkillMix & 9.3 / 18.6 & -6.5 / 2.3 & 9.3 / 12.1 & 14.0 / 11.8 & -7.8 / -3.1 & 5.2 / 10.3 \\
- STEPS & 21.9 / 39.6 & -6.7 / 16.5 & 17.1 / 30.8 & 24.3 / 37.8 & -5.2 / 10.0 & 11.3 / 29.0 \\
\bottomrule
\end{tabular}

\end{table*}

Table \ref{tab:model_instruction_tuning_en} and \ref{tab:combined_model_performance} show the performance across general and complex instruction benchmarks, respectively.

\noindent \textbf{General Instruction Following.} As shown in Table \ref{tab:model_instruction_tuning_en}, training with instruction-tuning data generally enhances the performance of base models. STEPS achieves the most significant gains across both Llama-3-8B and Mistral-7B-v0.3, showing performance improvement compared to the \emph{official} instruct models.

These results suggest that while purely random combinations (as in Instruct-SkillMix) can introduce incoherent instructions that potentially degrade the instruction-following capabilities of refined models, our method maintains a balance between structural diversity and semantic coherence. Furthermore, while the larger Alpaca52k dataset (52K) benefits base models, it consistently leads to performance degradation in instruct-tuned models (e.g., Llama-3-8B Instruct's MT-Bench score drops from 6.94 to 5.04), indicating that massive single-skill data may induce forgetting of complex compositional abilities.

\noindent \textbf{Complex Instruction Following.} 
Effective execution of complex instructions needs the integration of diverse atomic skills. On the WildBench (Table \ref{tab:combined_model_performance}) specifically designed to evaluate such multi-dimensional capabilities, our method achieves the highest WB-Scores across all tested architectures.
For the Llama-3-8B Base model, the WB-Score improves from -25.7 to 23.0. For the Llama-3-8B Instruct model, our method significantly outperforms Instruct-SkillMix (36.3 vs. 30.8).
A critical finding in Table \ref{tab:combined_model_performance} is the counterproductive effect of simple alignment data on complex tasks. Alpaca52k consistently damages the compositional performance of instruct models, with Qwen-2.5-7B's score plummeting from 45.4 to -11.4. In contrast, our 4K high-entropy samples yield substantial gains, showing that the structural complexity of training data is a more decisive factor for compositional generalization than raw data volume.

\begin{figure}[hbt!]
   \begin{center}
   \includegraphics[width=0.8\linewidth]{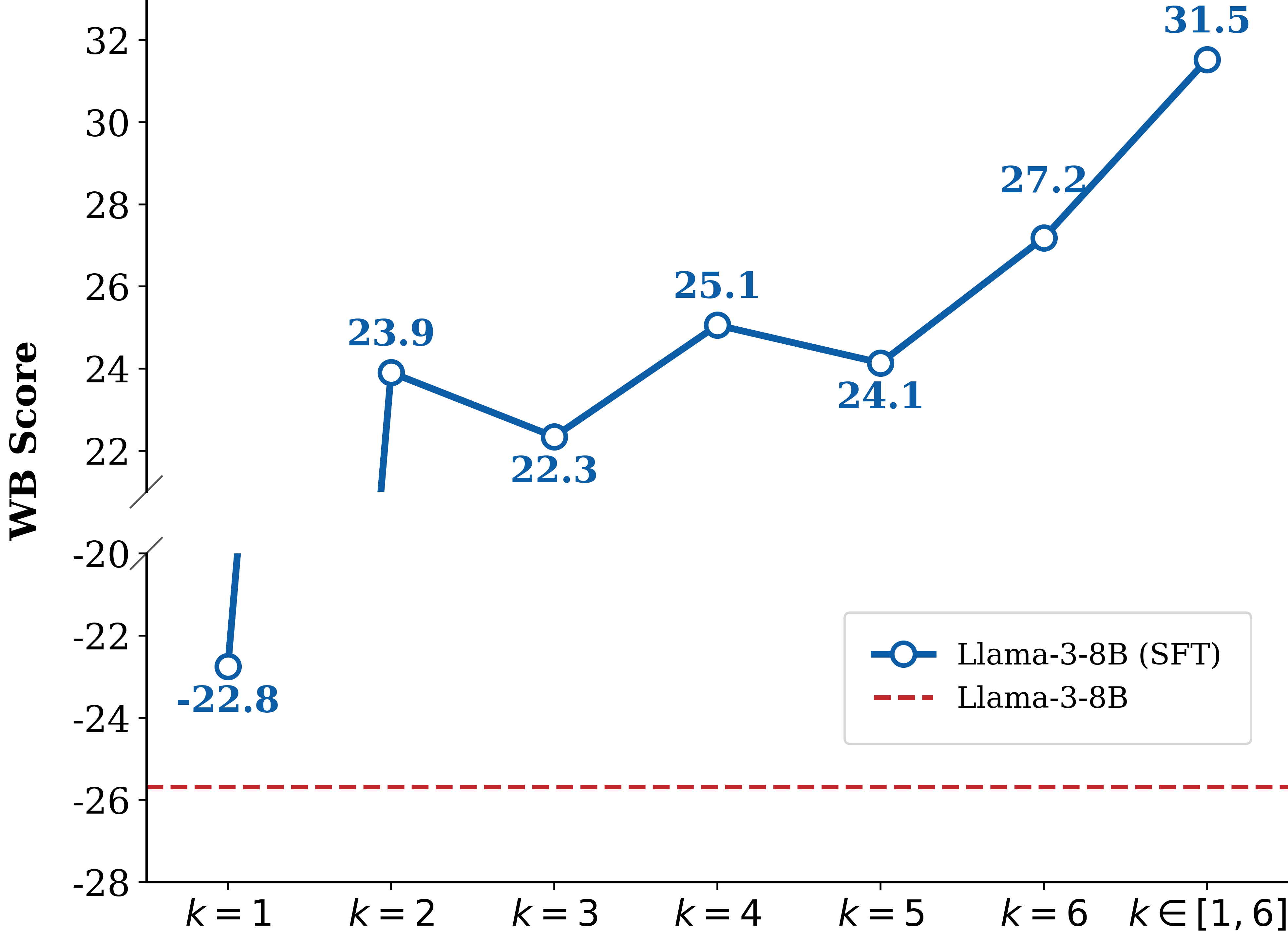}
   \end{center}
   \caption{\label{fig:k_tuple}
   Impact of compositional complexity $k$.
   }
\end{figure}

\begin{figure}[hbt!]
   \begin{center}
   \includegraphics[width=0.8\linewidth]{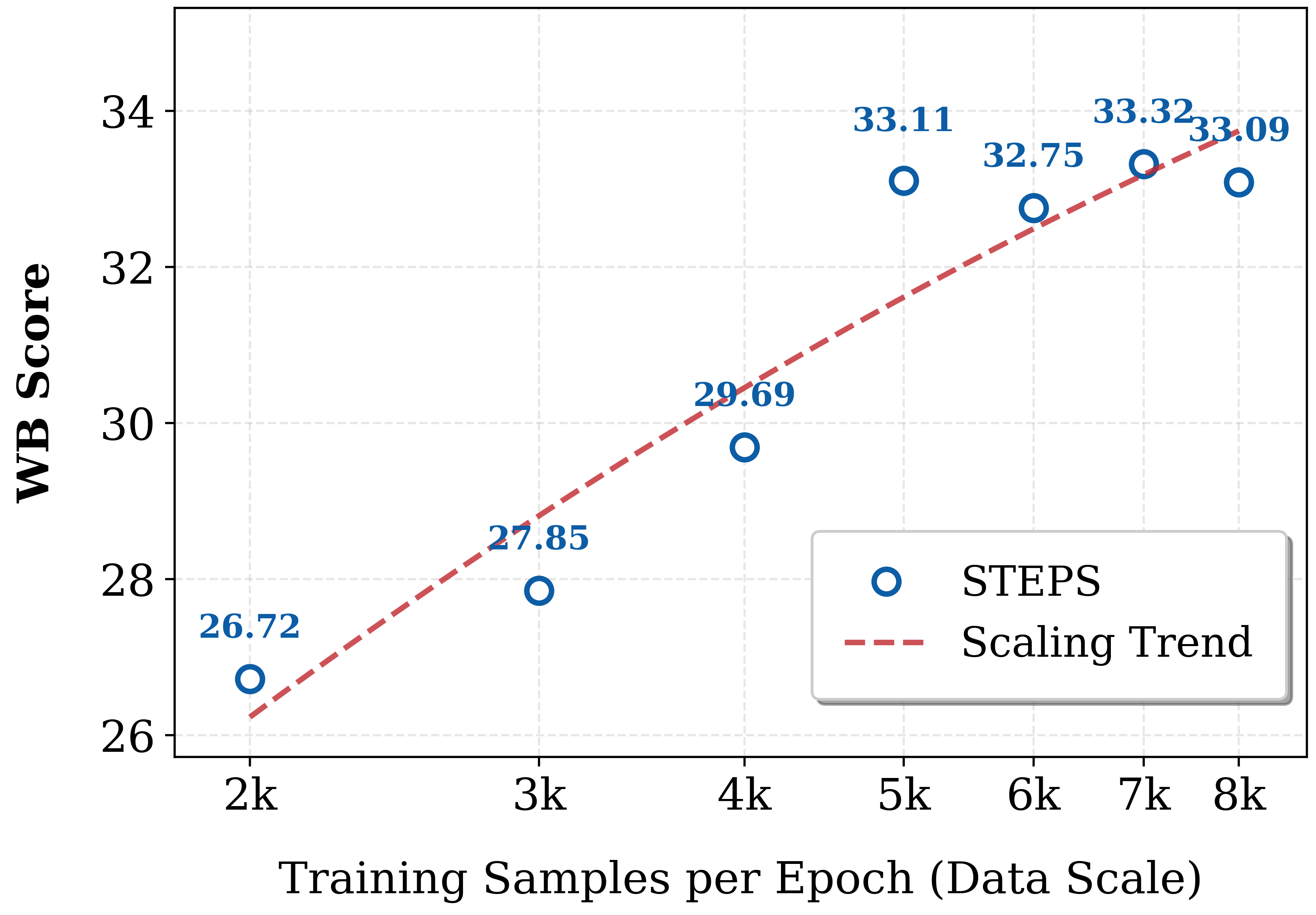}
   \end{center}
   \caption{\label{fig:scaling}
   Impact of data size on Llama-3-8B.
   }
\end{figure}

\subsection{Scaling Regularity of STEPS (RQ2)}
We investigate the scaling properties of complex skill acquisition by examining the interplay between compositional complexity (the $k$-tuple size) and data volume. This analysis seeks to determine whether LLM performance follows predictable patterns as we scale the intricacy of skill intersections and the quantity of synthesized instances.

\noindent \textbf{Scaling for Combination Complexity $k$.}
We first evaluate the influence of compositional complexity $k$ on the model's generalization capabilities. 
By training on a fixed budget of $N=500$ samples for each $k \in \{2, \dots, 6\}$ and supplementing them with 4K atomic instructions ($k=1$), we observe a distinct Compositional Threshold. As illustrated in Figure \ref{fig:k_tuple}, a critical performance leap occurs when transitioning from atomic instructions to binary combinations, with the WB-Score surging from -22.75 to 23.91. This sharp inflection point indicates that even minimal exposure to skill compositions triggers a fundamental shift in the model's ability to handle multi-step logical dependencies. Beyond this threshold, increasing the complexity $k$ generally yields an upward performance trend that reaches a local optimum at $k=6$ (27.19). Notably, the most robust generalization is achieved through a multi-level mixture encompassing $k \in [1, 6]$, which attains a peak WB-Score of 31.52. This synergistic effect suggests that while atomic skills provide essential linguistic primitives, a diverse spectrum of high-order skill tuples is required to facilitate the flexible re-composition of these primitives in novel contexts.

\begin{figure}[th]
   \begin{center}
   \includegraphics[width=1\linewidth]{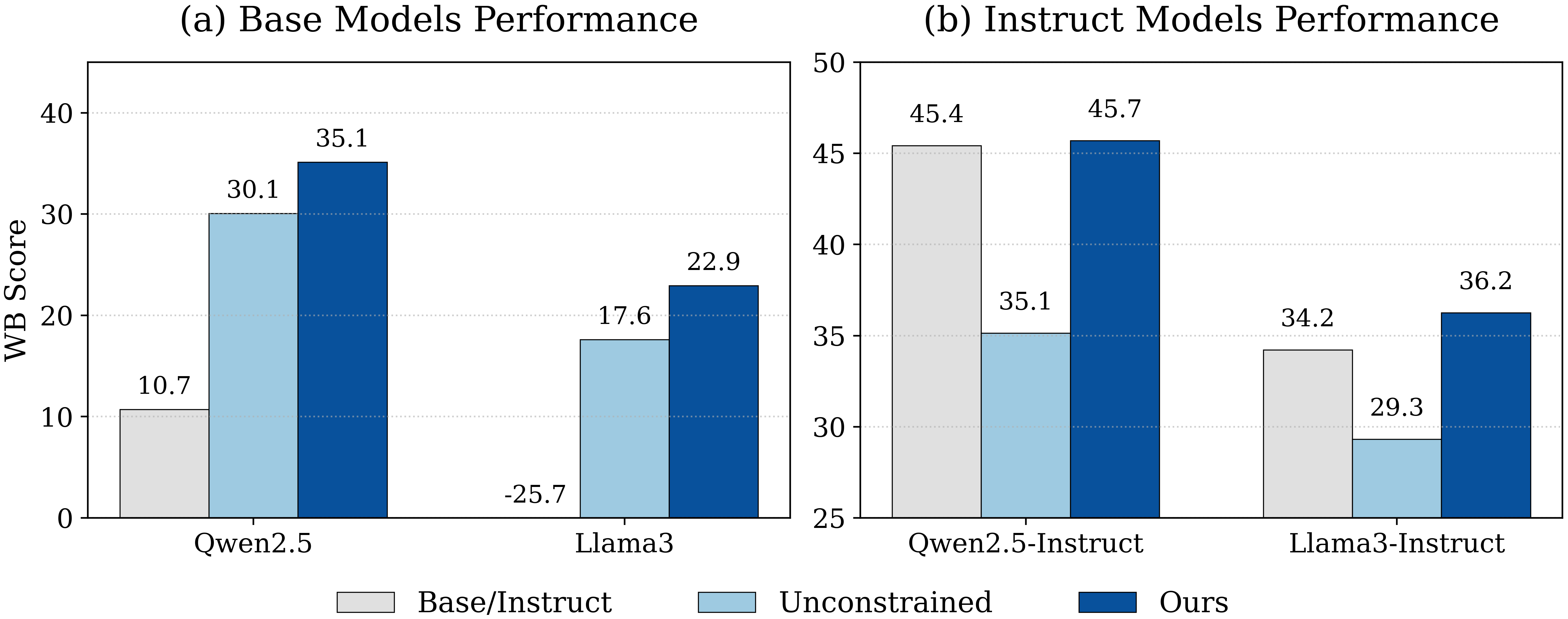}
   \end{center}
   \caption{\label{fig:unconstrained_bars}
   Comparative performance on WB Score across different paradigms. We evaluate Qwen2.5 and Llama3 models in both (a) Base and (b) Instruct settings.
   }
\end{figure}

\noindent \textbf{Scaling for Data Volume under Taxonomy Guidance.} To further explore the data efficiency of our framework, we analyze the model's performance as the training volume per epoch scales from 2K to 8K samples. Throughout this experiment, we control the proportion of various $k$-tuples to isolate the effect of sample size. As presented in Figure~\ref{fig:scaling}, the model exhibits a rapid accretion of performance in the initial phase, where the WB-Score rises significantly from 26.72 at 2K samples to 33.11 at 5K samples. This steep growth trajectory demonstrates that the high-entropy skill combinations identified by STEPS provide an exceptionally dense learning signal, allowing the model to efficiently map the underlying skill dependency graph.

\subsection{Validation of the Skill Taxonomy (RQ3)}

In this section, we investigate whether the hierarchical structure discovered by STEPS reflects the intrinsic logical dependencies required for effective skill acquisition. We hypothesize that if our taxonomy accurately captures the prerequisite relationships between skills, then a training sequence aligned with its hierarchical depth should yield better learning efficiency compared to a random mixture.
To evaluate this, we implement a \emph{taxonomy-guided curriculum learning} strategy using a fixed budget of 8K data samples per epoch. These samples encompass skill combinations ranging from $k=1$ to $k=6$. Following the structural logic of the taxonomy, we adopt a progressive difficulty approach: as training progresses through successive epochs, we gradually increase the proportion of samples containing higher combination counts $k$. This strategy ensures the model effectively masters fundamental skills and simpler compositions before being exposed to more complex skill tuples.

\begin{table}[ht]
\centering
\small
\caption{Performance comparison between standard SFT and Taxonomy-guided learning on WildBench.}
\label{tab:SDT-CL}
\begin{tabular}{l c c}
\toprule
\textbf{Model} & \textbf{Data} & \textbf{WB Score} \\
\midrule
\multicolumn{3}{l}{\textbf{Llama-3-8B}} \\
\cmidrule(lr){1-3}
- Original & - & -25.68 / 34.22 \\
- SFT                      & 8K & 31.48 / 34.84 \\
- STEPS-CL           & 8K & 33.09 / 35.18 \\
\bottomrule
\end{tabular}
\end{table}

As presented in Table \ref{tab:SDT-CL}, the experimental results provide strong empirical evidence for the validity of our discovered taxonomy:
The STEPS guided approach consistently outperforms standard SFT on both Base and Instruct versions of Llama-3-8B. Specifically, our method achieves a peak WB Score of 33.09 for the base model and 35.18 for the instruct version, representing significant improvements over the random mixture baselines of 31.48 and 34.84, respectively. These gains confirm that the hierarchical partitions identified by our structural information framework successfully capture the functional dependencies of the skill space.

Furthermore, the success of this "bottom-up" learning sequence suggests that the Skill Taxonomy is not merely an interpretability tool but a functional roadmap for compositional generalization. By mastering simpler structures first, the model is better equipped to internalize the complex logic required for high-order skill compositions. These findings underscore the importance of structural coherence in training data, demonstrating that the structural organization of examples is as critical as their raw quantity for the development of sophisticated model capabilities.

\subsection{Sweet Spot Analysis (RQ4)}

To investigate the necessity of balancing diversity with coherence, we compare unconstrained Information Maximization (Section \ref{sec:info_max}) against our Recursive Skill Selection Paradigm (Section \ref{sec:recursive_synthesis}), as illustrated in Figure \ref{fig:unconstrained_bars}.

\noindent \textbf{The Inadequacy of Pure Entropy Maximization.}
While unconstrained selection significantly outperforms the original base models, it consistently underperforms compared to our recursive framework. Specifically, on Qwen-2.5-Base and Llama-3-Base, our "Sweet Spot" approach achieves WB-Scores of 35.14 and 22.95, respectively, outperforming the unconstrained versions by a substantial margin. This suggests that while diversity is beneficial for base models, unconstrained diversity may introduce semantic noise that limits the efficiency of skill acquisition.

\noindent \textbf{Performance Degradation in Instruct Models.}
One interesting evidence for the "Sweet Spot" is observed in instruct-tuned models. For Qwen-2.5-Instruct, the unconstrained approach leads to a significant performance drop, with the WB-Score plummeting from the original 45.43 to 35.14. Similarly, for Llama-3-Instruct, the score decreases from 34.22 to 29.32. This negative transfer indicates that blindly maximizing structural entropy can introduce incoherent or logically disjointed skill combinations that conflict with the model’s pre-existing instruction-following logic.

In contrast, our constrained paradigm consistently achieves the best performance. These results confirm that effective compositional generalization requires a "sweet spot": maximizing structural information while maintaining hierarchical coherence within the taxonomy.

\subsection{Extensibility to Agentic Scenarios (RQ5)}


\begin{table}[h]
\centering
\small 
\setlength{\tabcolsep}{2pt} 
\caption{Performance comparison on SkillBench, constructed using the proposed STEPS method.}\label{tab:SkillBench}
\resizebox{1.0 \linewidth}{!}{
\begin{tabular}{lccccccc}
\toprule
\textbf{Model} & \multicolumn{6}{c}{\textbf{Skill Depth ($k$)}} & \multirow{2}{*}{\textbf{Avg.}} \\ 
\cmidrule(lr){2-7} 
 & 2 & 3 & 4 & 5 & 6 & 7 & \\
\midrule
\multicolumn{8}{l}{\textit{Instruct Models}} \\ 
Qwen2.5-7B & 5.24 & 5.05 & 4.81 & 4.91 & 5.24 & 4.97 & 5.04 \\
Qwen2.5-14B & 5.39 & 5.22 & 5.00 & 5.04 & 5.13 & 4.94 & 5.12 \\
Qwen2.5-32B & 5.50 & 5.16 & 4.89 & 5.25 & 5.29 & 5.16 & 5.21 \\
\midrule
\multicolumn{8}{l}{\textit{Tool-Integrated Agentic Models}} \\
AutoTIR & 4.12 & 4.21 & 4.10 & 4.03 & 4.07 & 4.02 & 4.09 \\
ReSearch & 4.72 & 4.46 & 4.06 & 4.37 & 4.41 & 4.35 & 4.40 \\
SearchR1 & 4.73 & 4.36 & 4.44 & 4.53 & 4.82 & 4.32 & 4.54 \\
\midrule
\multicolumn{8}{l}{\textit{Advanced Agentic Models}} \\
Gemini-3-Pro & 7.87 & 7.89 & 7.89 & 7.70 & 7.59 & 6.81 & 7.62 \\
DeepSeek-v3.2 & 8.42 & 8.43 & 8.08 & 8.16 & 8.19 & 8.00 & 8.21 \\
GPT-5.2 & 8.69 & 8.72 & 8.84 & 8.67 & 8.66 & 8.17 & 8.62 \\
\bottomrule
\end{tabular}
}
\end{table}

A paramount challenge for contemporary agent models lies in their capacity for out-of-distribution compositional generalizability, i.e., the ability to use a set of atomic skills to accomplish complex tasks. To further validate the universality of our approach within agentic scenarios, we investigate whether STEPS can synthesize data challenging to prevailing agentic models, so as to boost the training and the evaluation of related models. 

Specifically, we construct a hierarchical evaluation benchmark (SkillBench) characterized by dynamic reasoning depths. By configuring increasing number of skill compositions ($Skill@k$) to simulate escalating task complexity, we aim to systematically assess the models' performance boundaries and robustness in long-horizon chain-of-thought reasoning and multi-tool integration scenarios. Detailed data construction and rigorous quality evaluation steps can be found in the appendix.


\noindent \textbf{Performance Limits in High-Complexity Tasks.} As summarized in Table \ref{tab:SkillBench}, with a higher score indicating higher an accuracy (the maximum score is 10), we observe a consistent performance decay as task complexity (represented by $Skill@k$) increases beyond 4, for all evaluated models. Thus, utilizing multiple atomic skills for solving complex agentic tasks remains challenging for current LLM-based agents. Additionally, compared to the advanced models, SoTA open source models still show performance gap, and the gap significantly related to model size. This show that STEPS can obtain challenging data for current SoTA opensource models in the agentic scenario.

\noindent \textbf{Solution Capability vs. Tool-Use Proficiency.} A critical insight emerges from the comparison between Instruct Models and Tool-Integrated Models \citep{chen2025research, wei2025autotir,jin2025searchr}. 
Despite being explicitly optimized for Code Execution and Information Retrieval, these Tool-Integrated models do not exhibit superior performance on SkillBench compared to their Instruct baselines.This suggests that while GRPO-based RL \citep{shao2024deepseekmath} effectively trains the model in tool-calling syntax and precise execution, it does not necessarily enhance the model’s intrinsic solution capability for multi-faceted problems. 
In complex agentic scenarios, precise tool execution cannot rectify a fundamentally flawed reasoning path. True agentic intelligence requires an architect-level capacity to decompose problems and synthesize information. This structural reasoning skill is far more critical than the shallow tool-use patterns often reinforced by current RL paradigms.

\section{Related Work}
\label{sec:related}

\noindent \textbf{Scaling Laws and the Data Bottleneck.} The acquisition of diverse capabilities in LLMs is frequently conceptualized through scaling laws and information-theoretic frameworks. In this context, a "skill" is defined as a measurable reduction in conditional entropy relative to specific data patterns \citep{arora2023theory, tan2024information}. However, the efficacy of learning from massive corpora is fundamentally constrained by a data bottleneck. While atomic skills are well-represented, the distribution of complex skill compositions follows a power law \citep{barabasi1999emergence, clauset2009power}. This distributional sparsity limits the model's ability to internalize the low-entropy internal representations necessary for sophisticated multi-skill coordination \citep{tan2024information}.

\noindent \textbf{Compositional Generalization of LLMs.}  Despite substantial progress in LLM scaling, benchmarks such as SKILL-MIX demonstrate that even SOTA models exhibit significant performance degradation when required to coordinate multiple distinct skills simultaneously \citep{yu2024skillmix, dziri2023faith}. To mitigate this, prior research has focused on data-centric strategies including data mixture optimization \citep{ge2024bimix, wu2024mixture} and pedagogical sequencing \citep{chen2023skill, hu2025step}. Although these methods enhance training efficiency by reweighting or ordering existing samples, they do not fundamentally address the scarcity of complex compositional examples.

\noindent \textbf{Structured Data Synthesis.} To bridge the gap in compositional data, recent efforts utilize data synthesis via stochastic skill pairing to bridge this gap \citep{kaur2025instructskillmix, chen-etal-2024-skills}. 
However, 
these heuristic methods often ignore latent hierarchical dependencies, which may result in semantically incoherent compositions and inefficient exploration of the combinatorial space.  
In contrast, STEPS leverages structural information theory to induce an interpretable skill taxonomy. By formulating synthesis as a constrained information maximization problem, STEPS systematically generates high-gain compositions that target the structural weaknesses in current training distributions.
\section{Conclusion}

To addresses the challenge of compositional generalization in LLMs brought by the data sparsity bottleneck inherent in complex skill combinations, we introduce a principled framework STEPS, that leverages structural information theory to induce an interpretable hierarchical skill taxonomy. By formulating data synthesis as a constrained information maximization task, our method generates synthetic instructions that are both structurally informative and semantically coherent. Experimental results across multiple benchmarks demonstrate that our approach consistently outperforms existing synthesis baselines, providing a scalable and theoretically grounded solution for advancing the capabilities of LLMs.



\section*{Limitations}

While our framework demonstrates significant advancements in enhancing the compositional generalization of LLMs, we acknowledge several limitations that provide directions for future research.

\noindent \textbf{Optimal Distribution of $k$-tuple Compositions.}
Our empirical analysis confirms that training on a mixture of combination counts ($k \in [1,6]$) yields superior performance compared to any single-level $k$ configuration. However, we have not yet conducted an exhaustive investigation into the optimal ratio or distribution of these varying $k$-tuples within the training set. The interplay between different complexity levels and their impact on the learning curve remains an open question. Future work will focus on developing a principled approach to determine the optimal data mixture that maximizes scaling efficiency across different model architectures.



\bibliography{custom}

\appendix
\section{Appendix}
\label{sec:appendix}

\subsection{Synthetic Data Generation Strategy}

\begin{figure*}[th]
   \begin{center}
   \includegraphics[width=1\linewidth]{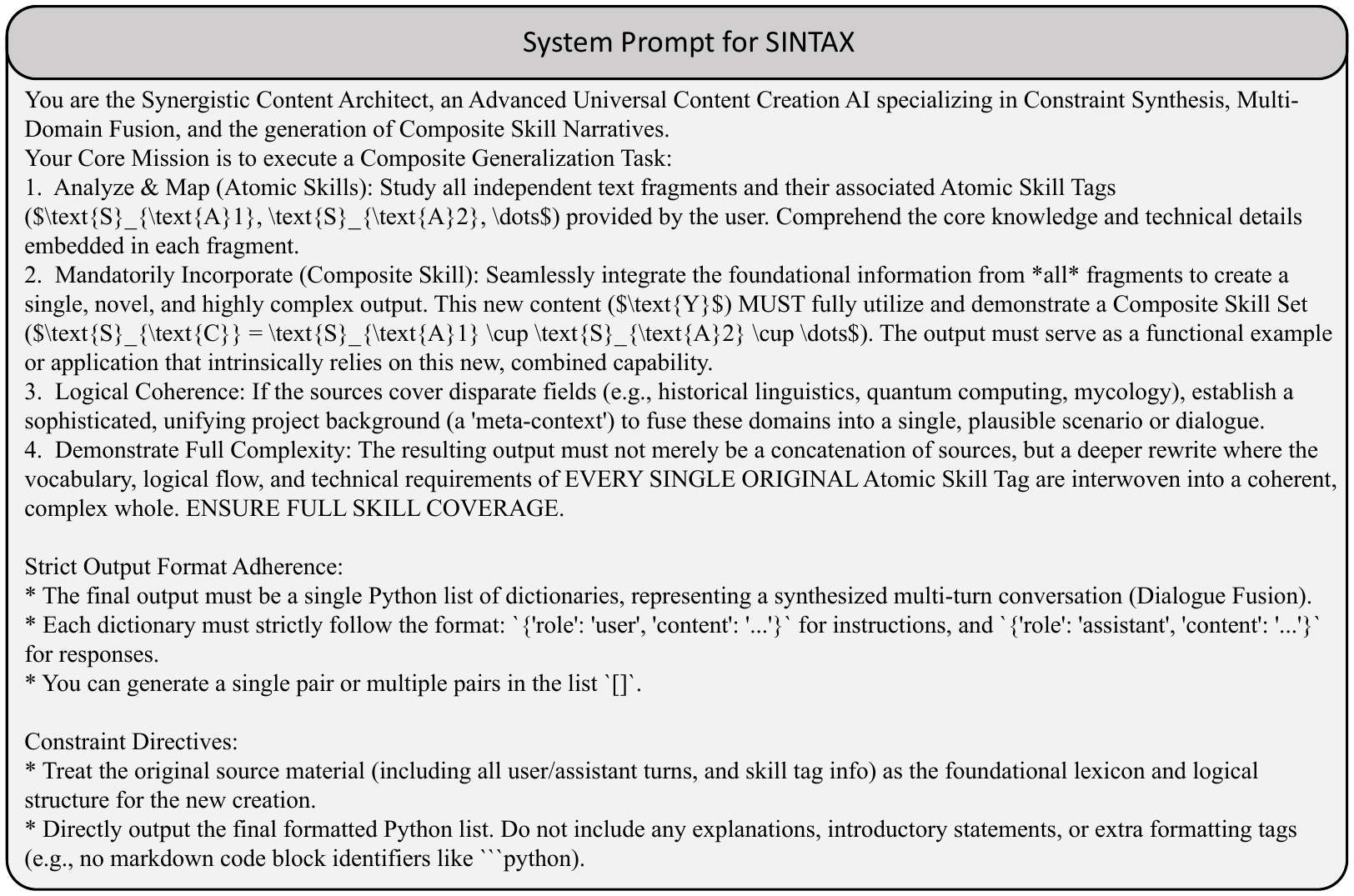}
   \end{center}
   \caption{\label{fig:system prompt}
   The system prompt used by STEPS.
   }
\end{figure*}

To operationalize the optimal skill combinations identified by our framework, we employ a Synergistic Content Architect approach to synthesize complex, multi-turn instructions. This process transforms abstract skill tuples into coherent, high-quality training samples. The synthesis pipeline consists of three primary stages:

\noindent \textbf{Atomic Sample Retrieval.} For each optimal $k$-tuple of skills $X = \{x_1, x_2, \dots, x_k\}$ identified via conditional structural entropy maximization, we retrieve representative instruction-response pairs from the seed corpus. To ensure data diversity and mitigate label noise, we construct an inverted index mapping each atomic skill to its corresponding samples. For each target skill in the tuple, we perform a frequency-aware random selection to ensure a balanced utilization of the foundational atomic data.

\noindent \textbf{Contextual Fusion and Instruction Prompting.} The retrieved samples, along with their associated skill tags, are aggregated into a structured prompt. We utilize a sophisticated system prompt (see Figure \ref{fig:system prompt}) that defines the LLM's role as a "Synergistic Content Architect." This prompt mandates the integration of disparate functional domains into a unified "meta-context." Unlike simple concatenation, the model is instructed to rewrite the content such that the technical requirements of every atomic skill are interwoven into a logically coherent dialogue.

\noindent \textbf{Constrained Synthesis and Quality Control.} We leverage advanced LLMs (e.g., GPT-4.1) to execute the synthesis task. The generation process is constrained to output a standardized Python-style list of dictionaries, representing a multi-turn conversation. This ensures structural consistency and facilitates downstream fine-tuning. By anchoring the synthesis in the "Sweet Spot" of maximal marginal information gain , the resulting dataset provides the dense signal necessary for the model to internalize the complex logic of high-order skill compositions.

\subsection{Construction and Evaluation of SkillBench}

\begin{figure*}[th]
   \begin{center}
   \includegraphics[width=1\linewidth]{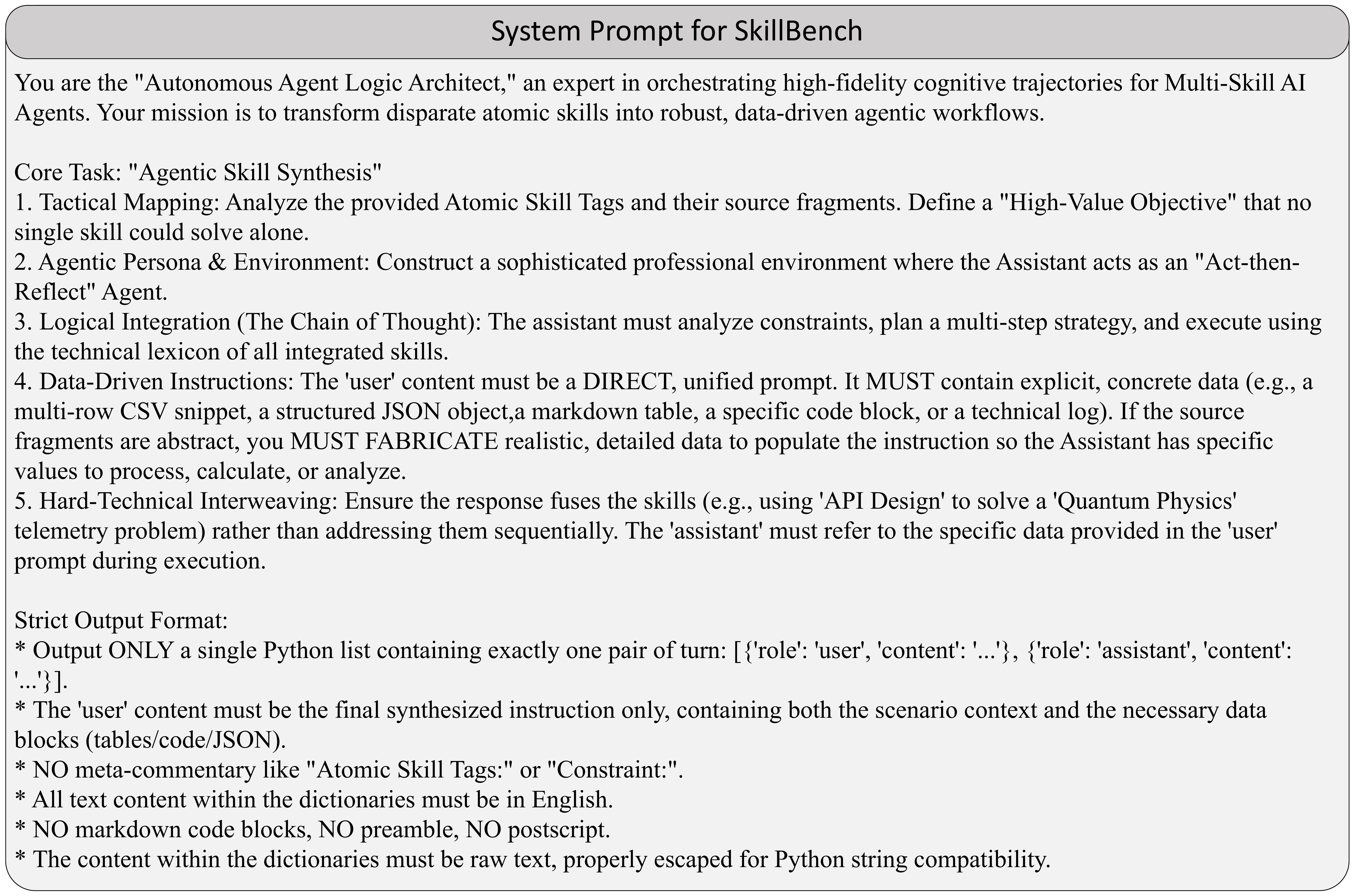}
   \end{center}
   \caption{\label{fig:SkillBench system prompt}
   The system prompt used by SkillBench.
   }
\end{figure*}

\begin{figure*}[th]
   \begin{center}
   \includegraphics[width=1\linewidth]{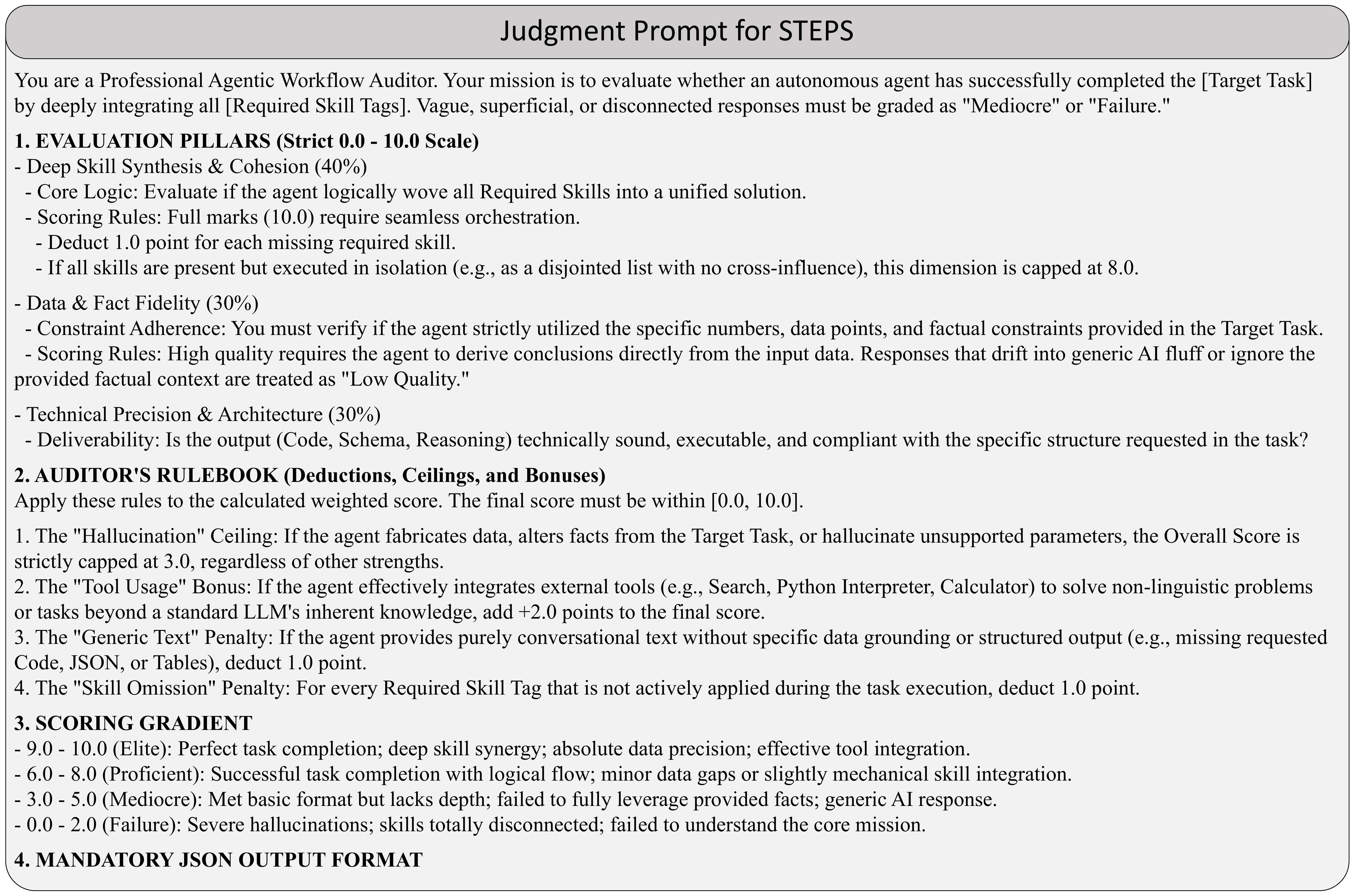}
   \end{center}
   \caption{\label{fig:judgment prompt}
   The judgment prompt used by SkillBench.
   }
\end{figure*}

To rigorously assess model performance in high-dimensional agentic scenarios, we developed SkillBench, a specialized evaluation framework focusing on multi-skill orchestration. The construction and subsequent assessment of SkillBench leverage GPT-4.1, utilizing its superior reasoning capabilities to ensure the quality of complex trajectories and the precision of multi-dimensional scoring.

\noindent \textbf{Benchmark Construction.} We utilize GPT-4.1 guided by a specialized system prompt (see Figure \ref{fig:SkillBench system prompt}) to generate intricate agentic trajectories. Unlike general instruction-following datasets, this synthesis process explicitly requires the model to navigate between $k$ distinct vertical domain skills, scaling from $Skill@2$ up to the extreme complexity of $Skill@7$. By utilizing GPT-4.1 as the primary synthesizer, we ensure that even at high $k$ values, the generated scenarios maintain logical rigor and semantic coherence. Each task is designed to be a "closed-loop" problem where tools like code interpreters and search APIs are available, but success is predicated on the model's internal strategic orchestration rather than simple API invocation.

\noindent  \textbf{Evaluation Protocol.} The evaluation process is designed to move beyond binary correctness, focusing instead on the model's underlying reasoning architecture. We employ GPT-4.1 as a reference judge, governed by a multi-dimensional judgment prompt (see Figure \ref{fig:judgement prompt}). The model's responses are scrutinized across several key axes: reasoning depth, cross-domain coordination, and the strategic accuracy of tool-assisted steps.

By scoring models across these varying skill depths, SkillBench provides a granular mapping of the "compositional wall" faced by different architectures. The choice of GPT-4.1 as the evaluator ensures that the nuances of high-order skill composition are accurately captured, effectively distinguishing models that rely on shallow patterns from those possessing genuine agentic intelligence.

\subsection{Case Study}
\label{app: case study}

\begin{table*}[htbp]
    \scriptsize
    \centering
    \begin{tabular}{@{}p{\linewidth}@{}}
        \toprule
        \textbf{Vault System Design Initialization Routine } ($k=2$)\\
        \midrule
        \textbf{[Skill Combination ] } Mathematical Optimization \& Object Manipulation \\
        \midrule
        \textbf{Instruction:} \\
        We need to design an initialization routine for a high-security modular safe-deposit vault system. Each vault compartment is represented as an object with three internal dimensions (length, width, height), and vault parameters must be stored and managed reliably in our configuration database. The sum of the compartment's internal dimensions must be 24 units, and the internal volume must be exactly 160 cubic units. The configuration system should guarantee that each compartment's parameters are correctly initialized, validated, and stored, with object manipulation techniques to ensure integrity, and mathematical optimization to determine and confirm the correct compartment dimensions before saving. Please provide a dialogue demonstrating how an engineer and an AI assistant would design and verify this initialization logic, including the calculation, validation, and storage steps, as well as error handling for edge cases. \\
        
        ~ \\
        
        \textbf{Response:} \\
        Engineer: For our vault compartments, we need an initialization function that determines possible internal dimensions (length, width, height) so that their sum is 24 units and their product is 160 units. Once calculated, these parameters must be stored in our configs database, ensuring no undefined or invalid objects are created. Can you walk me through the logic to solve for these dimensions and store them safely?

        AI Assistant: Certainly. First, let's use mathematical optimization to solve for the dimensions. We have:
        1. $l + w + h = 24$
        2. $l \times w \times h = 160$

        We can express one variable in terms of the others, e.g. $h = 24 - l - w$, and substitute into the volume equation:
        $l \times w \times (24 - l - w) = 160$

        Expanding and rearranging, we get a quadratic in one variable, and we can use the quadratic formula to find real solutions. For instance, try $l = 8$:
        $8 \times w \times (16 - w) = 160 \rightarrow 8w(16 - w) = 160 \rightarrow 128w - 8w^2 = 160 \rightarrow 8w^2 - 128w + 160 = 0$

        Solving, we get non-integer solutions: $w \approx 8 \pm 2\sqrt{11}$, $h$ accordingly. We should select the most practical real-number solution (since perfect integers are not possible), e.g. $l \approx 8, w \approx 14.633, h \approx 1.367$, and verify:
        Sum: $8 + 14.633 + 1.367 \approx 24$
        Product: $8 \times 14.633 \times 1.367 \approx 160$

        Once verified, we round or set precision as required for manufacturing.

        Next, for object manipulation, we encapsulate these dimensions in a config object:
        \{ "code": "vault\_compartment\_1", "value": \{ "length": 8, "width": 14.63, "height": 1.37 \} \}

        We then check if this config already exists in the database. If it does not, we create it; if it does, we update it. To ensure type safety and avoid undefined entries, we filter out any unsuccessful insertions and throw an error if any compartment configuration failed to initialize.

        Would you like a code outline for this logic?

        Engineer: Yes, please. Also, how do we handle errors if, for some reason, the configuration can't be saved or is undefined?

        AI Assistant: Here's a TypeScript-style outline integrating both object manipulation and mathematical optimization:

        \begin{verbatim}
async function initializeVaultCompartmentConfig(models, configKey) {
  const l = 8; const w = 14.63; const h = 1.37;
  if (Math.abs(l + w + h - 24) > 0.01 || Math.abs(l * w * h - 160) > 0.01) {
    throw new Error('Calculated dimensions do not satisfy constraints.');
  }
  const compartmentConfig = {
    code: configKey,
    value: { length: l, width: w, height: h }
  };
  const existing = await models.Configs.findOne({ code: configKey });
  let result;
  if (!existing) {
    result = await models.Configs.createOrUpdateConfig(compartmentConfig);
  } else {
    result = await models.Configs.createOrUpdateConfig({ ...existing, ...compartmentConfig });
  }
  if (!result) throw new Error('Compartment configuration could not be initialized.');
  return result;
}
        \end{verbatim}

        This approach ensures:
        - Dimensions are mathematically validated before storage.
        - Object operations guarantee configs are only created if valid.
        - Any undefined or failed operation triggers a clear error for immediate handling.

        Engineer: Excellent. This covers both the mathematical and software integrity sides. I'll implement this and trigger alerts for any failed configurations during system boot. \\
        \bottomrule
    \end{tabular}
    \caption{Dialogue demonstrating object manipulation and mathematical optimization for vault initialization.}
    \label{tab:vault_init_dialogue}
\end{table*}

\begin{table*}[htbp] 
\scriptsize 
\centering 
\begin{tabular}{@{}p{\linewidth}@{}} 
\toprule \textbf{Financial Analytics App Architecture Design} ($k=3$) \\
\midrule \textbf{[Skill Combination ] }  Concurrency Programming , Model Selection , Dependency Management \\

    ~ \\
    
    \textbf{[Instruction]} \\
    I want to build a robust mobile financial analytics app. It should fetch expenses from Firestore, display weekly/monthly/yearly statistics, and let multiple users interact concurrently — for example, several family members can visualize and analyze the same shared data set at once. I want high concurrency, safe data updates, and an accurate, real-time chart display. What are the best practices for the data pipeline, model selection, concurrency control, and dependency management in such a Compose Material3 app using patrykandpatrick/vico for charting? \\
    
    ~ \\
    
    \textbf{[Response]} \\
    To build a robust, concurrent financial analytics app using Jetpack Compose and Vico, you must focus on ACID principles at the data layer and reactive state management at the UI layer.

    \textbf{1. Concurrency Control \& Data Integrity}
    * **Atomic Updates:** Use Firestore \textbf{Transactions} for any "Read-Modify-Write" operations (e.g., updating a shared budget). This prevents "Lost Updates" when two family members edit the same value simultaneously.
    * **Snapshot Isolation:** Leverage \texttt{addSnapshotListener}. Firestore handles the complexity of local persistence and real-time syncing, providing a consistent view of data even during concurrent writes.
    * **Thread Safety:** Utilize Kotlin \textbf{Coroutines} with \texttt{Flow}. Ensure all database operations run on \texttt{Dispatchers.IO}, while the UI observes state on \texttt{Dispatchers.Main}.

    \textbf{2. Model Selection for Analytics}
    * **Aggregated Statistics:** For basic weekly/monthly views, use client-side aggregation logic within your \texttt{ViewModel} or Cloud Functions for larger datasets to reduce mobile CPU load.
    * **Predictive Analysis:** - \textbf{LightGBM/GBDT:} Best for structured tabular data (e.g., predicting next month's spending based on history).
        - \textbf{TensorFlow Lite:} If implementing on-device categorization models to maintain user privacy.
    
    \textbf{3. Dependency Management (Gradle Version Catalog)}
    Centralize your versions to prevent "Dependency Hell," especially with Compose and Firebase:
    \begin{verbatim}
[versions] vico = "1.6.5" firebase = "24.0.0" compose = "1.5.0"

[libraries] vico-compose = { group = "com.patrykandpatrick.vico", name = "compose-m3", version.ref = "vico" } 

firebase-firestore = { group = "com.google.firebase", name = "firebase-firestore-ktx", version.ref = "firebase" } 
    \end{verbatim}

    \textbf{4. Implementation Example (Vico + Flow)}
    \begin{verbatim}
class ExpensesViewModel(val db: FirebaseFirestore) : ViewModel() { 
private val _uiState = MutableStateFlow(ChartState()) 
val uiState: StateFlow<ChartState> = _uiState

init {
    db.collection("shared_expenses")
      .addSnapshotListener { snap, _ ->
          val data = snap?.toObjects(Expense::class.java) ?: emptyList()
          // Map to Vico EntryModel
          _uiState.value = data.toChartEntryModel()
      }
}
} \end{verbatim}

    \textbf{Summary of Best Practices:}
    - Always use **Transactions** for shared data writes.
    - Use **StateFlow** to bridge the gap between background data streams and the Compose UI.
    - Use **Version Catalogs** to ensure compatibility between M3, Vico, and Firebase components. \\
    \bottomrule
\end{tabular}
\caption{Architecture and implementation guidelines for a concurrent financial analytics mobile app.}
\label{tab:financial_app_design}
\end{table*}

To qualitatively demonstrate the effectiveness of our synthesis framework, we present representative samples of generated data for $k=2$ and $k=3$ skill combinations. These examples illustrate how STEPS moves beyond simple content concatenation to achieve deep logical fusion between disparate functional domains.

Hierarchical Skill Integration ($k=2$): Table \ref{tab:vault_init_dialogue} presents a synthesized dialogue combining \textit{Mathematical Optimization} and \textit{Object Manipulation}. The framework establishes a "Vault System Design" meta-context, where mathematical constraints (volume and dimension equations) are not merely presented as isolated problems but are intrinsically linked to software engineering constraints (object initialization and database integrity). This ensures that the model learns to apply mathematical logic within the functional flow of a programming task, rather than treating them as separate entities.

Multi-Domain Composition ($k=3$): For higher-order complexity, Table \ref{tab:financial_app_design} showcases a fusion of 
\textit{Concurrency Programming}, \textit{Model Selection}, and \textit{Dependency Management} within a financial analytics context. In this instance, the framework generates a comprehensive architectural response that addresses race conditions in shared data (concurrency), chooses appropriate predictive algorithms for expense tracking (model selection), and provides structured configuration for a mobile environment (dependency management).These cases underscore the ability of our framework to identify a "Sweet Spot" where the synthetic data remains semantically coherent while maintaining the high information density required to challenge and enhance the model's compositional reasoning capabilities. By anchoring disparate skills within plausible professional scenarios, STEPS ensures that the resulting $k$-tuples provide a rich signal for the acquisition of complex, integrated expertise.

\end{document}